\documentclass[10pt,twocolumn,letterpaper]{article}

\usepackage[accsupp]{axessibility}
\usepackage[pagenumbers]{cvpr} 

\definecolor{cvprblue}{rgb}{0.21,0.49,0.74}
\usepackage[pagebackref,breaklinks,colorlinks,allcolors=cvprblue]{hyperref}

\usepackage{algorithm}
\usepackage{algorithmic}
\usepackage{pifont} 
\usepackage{tabularx}
\usepackage{graphicx}
\usepackage{booktabs}
\usepackage[table]{xcolor}
\usepackage{colortbl}
\usepackage{amsmath,amsfonts,dsfont}
\usepackage{amsmath}
\usepackage{multirow}

\def\paperID{****}
\def\confName{CVPR}
\def\confYear{2026}

\title{All-in-One Slider for Attribute Manipulation in Diffusion Models}

\newcommand*\samethanks[1][\value{footnote}]{\footnotemark[#1]}

\author{Weixin Ye$^{1,2}$\thanks{Equal contribution: weixinye@bjtu.edu.cn, hgzhu@cityu.edu.mo} \ \ \  Hongguang Zhu$^{3}$\samethanks \ \ \  Wei Wang$^{1,2}$\thanks{Corresponding author: wei.wang@bjtu.edu.cn} \ \ \  Yahui Liu$^{4}$ \ \ \  Mengyu Wang$^{1,2}$ \ \ \  Xuecheng Nie$^{5}$ \\
$^1$Institute of Information Science, Beijing Jiaotong University \\
$^2$Visual Intelligence + X International Cooperation Joint Laboratory of the Ministry of Education \\
$^3$City University of Macau \quad
$^4$Kuaishou \quad
$^5$Meitu \\
}

\begin{document}
\maketitle
\begin{abstract}
Text-to-image (T2I) diffusion models have made significant strides in generating high-quality images. However, progressively manipulating certain attributes of generated images to meet the desired user expectations remains challenging, particularly for content with rich details, such as human faces. Some studies have attempted to address this by training slider modules. However, they follow a \textbf{One-for-One} manner, where an independent slider is trained for each attribute, requiring additional training whenever a new attribute is introduced.
This not only results in parameter redundancy accumulated by sliders but also restricts the flexibility of practical applications and the scalability of attribute manipulation. To address this issue, we introduce the \textbf{All-in-One} Slider, a lightweight module that decomposes the text embedding space into sparse, semantically meaningful attribute directions. Once trained, it functions as a general-purpose slider, enabling interpretable and fine-grained continuous control over various attributes. Moreover, by recombining the learned directions, the All-in-One Slider supports the composition of multiple attributes and zero-shot manipulation of unseen attributes (e.g., races and celebrities). Extensive experiments demonstrate that our method enables accurate and scalable attribute manipulation, achieving notable improvements compared to previous methods. Furthermore, our method can be extended to integrate with the inversion framework to perform attribute manipulation on real images, broadening its applicability to various real-world scenarios. 
The code is available on our \href{https://github.com/ywxsuperstar/ksaedit}{project page}.

\end{abstract}    
\section{Introduction}
\label{sec:intro}
\begin{figure}[t]
    \centering
    \includegraphics[width=1.0\linewidth]{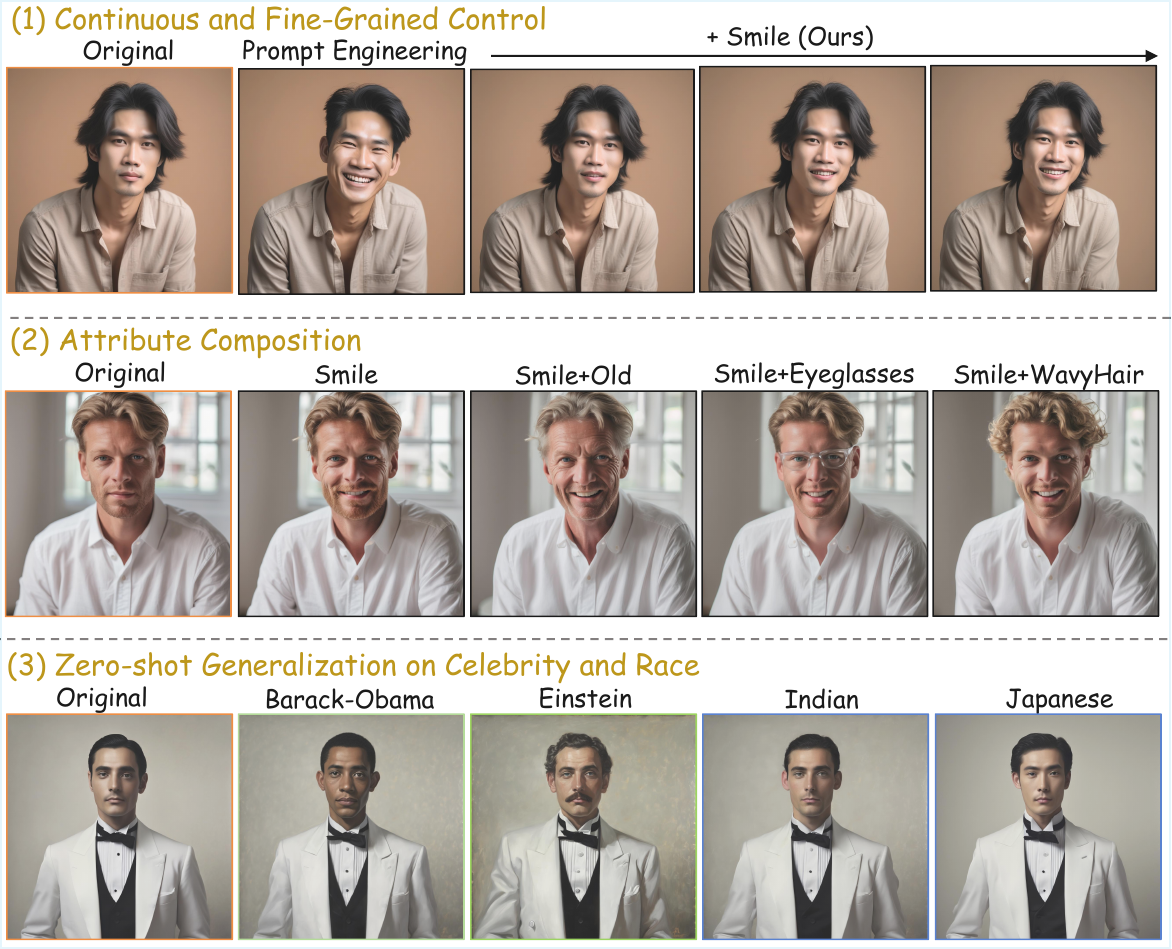} 
    \vspace{-6.6mm}
    \caption{Our All-in-One Slider shows advantages in: (1) Fine-grained and continuous control over desired attribute, without affecting other attributes (e.g., subject identity and appearance).
    (2) Combination of multiple facial attributes (e.g., smile and age) for consistent and conflict-free transformations.
    (3) Zero-shot generalization to unseen attributes, without multiple and cumbersome training processes.} 
    \vspace{-6mm}
    \label{fig:intro-visuli} 
\end{figure}

With the widespread emergence of diffusion models~\cite{wu2023uncovering,podell2023sdxl}, text-to-image (T2I) generation has advanced significantly in recent years. Despite their impressive performance, a major challenge remains: providing users with continuous and fine-grained control functions over specific image attributes, which is essential for users to adjust generated images to obtain the desired results. As illustrated by the example in Figure~\ref{fig:intro-visuli}(1), traditional prompt engineering approaches, such as simply appending phrases like ``\textit{with a big smile}'' to the base prompt, ``\textit{A photo of a gentle Thai man with layered hair, in a warm-toned studio}'', often result in coarse-grained and rigid manipulations. Such textual variations not only prevent users from specifying the intensity of the smile but also inadvertently affect many unrelated attributes, such as the subject's hairstyle and identity, leading to suboptimal outcomes.

To enable attribute manipulation, early studies focused on exploring the representation properties of generative models and identifying the semantic directions corresponding to attributes.
These efforts have spanned different representation spaces, including the \textit{UNet bottleneck layer}~\cite{kwon2022diffusion,park2023understanding}, \textit{text embedding space}~\cite{gandikota2024concept,baumann2025continuous,hertz2022prompt}, \textit{noise space}~\cite{dalva2024noiseclr}, \textit{weight space}~\cite{dravid2024interpreting,gandikota2024concept}, \textit{representation in GAN-based architecture}~\cite{li2024stylegan,tov2021designing}, and more recently, rectified flow trajectories~\cite{Dalva_2025_CVPR,rout2025semantic}.
However, these methods often suffer from semantic entanglement~\cite{kwon2022diffusion,park2023understanding}, lack clear attribute correspondence~\cite{dalva2024noiseclr}, or rely on the addition of phrases with limited continuity of attribute manipulation~\cite{hertz2022prompt}. Some other studies~\cite{li2024stylegan,tov2021designing} achieve precise control, but their methods are designed for the GAN spaces, restricting the generalization to the powerful diffusion models. As a result, these methods fail to achieve both fine-grained and continuous attribute manipulation within a single framework.

\begin{figure}[!t]
    \centering
    \includegraphics[width=0.9\linewidth]{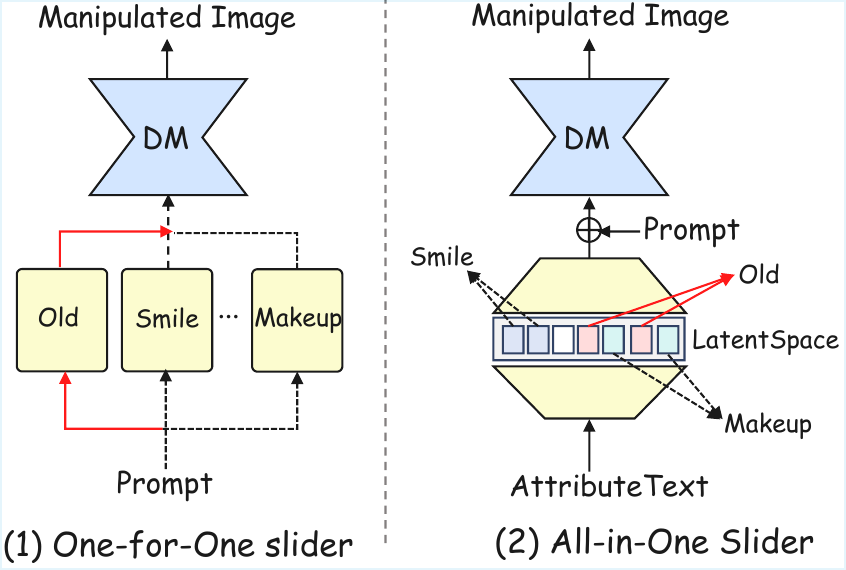}
    \vspace{-3mm}
    \caption{(1) Existing One-for-One slider methods require training a specific slider module for each attribute.
    (2) Our All-in-One slider only needs training once to obtain a unified and disentangled latent space for various attributes, supporting the flexible manipulation of multiple diverse attributes.
    } 
    \vspace{-5mm}
    \label{fig:diff_flow} 
\end{figure}

Recent studies~\cite{dravid2024interpreting,gandikota2024concept,baumann2025continuous,attadapter2025} have made efforts in achieving continuous attribute manipulation (\textit{e.g.}, gradually intensifying one’s smile or progressively increasing the degree of aging). Some methods ~\cite{dravid2024interpreting,gandikota2024concept,attadapter2025} rely on domain-specific LoRA or adapter training to achieve latent space disentanglement.
Others~\cite{baumann2025continuous} introduce attribute-wise training, which enables continuous attribute manipulation but requires attribute-wise supervision and heavily relies on labeled or paired data.
As a result, these methods typically require separate training for each attribute and follow a \textbf{One-for-One} design paradigm. In their pipeline, each attribute is handled in isolation via a dedicated slider module or a specific optimized representation direction, as shown in Figure~\ref{fig:diff_flow}(1).
When manipulating multiple attributes, this One-for-One paradigm requires multiple training runs, leading to significant time and computational costs and limiting their scalability and applicability in real-world scenarios.

Facing the limitations of the One-for-One paradigm, we propose an \textbf{All-in-One Slider} framework, as illustrated in Figure~\ref{fig:diff_flow}(2), which uses a single lightweight module to achieve flexible control over multiple attributes. 
In order to achieve generalized control of multiple attributes within a single module, a key challenge lies in disentangling these attributes. Once disentangled, the intensity of a target attribute can be easily controlled without affecting others, therefore enabling generalized and precise attribute manipulation.
Inspired by the recent studies of sparse autoencoder in LLMs~\cite{karvonen2024measuring,huben2023sparse,zhao2025sparse}, we adopt the idea of ``\textit{break it down to build it up}'' and propose an \textbf{Attribute Sparse Autoencoder}, which is implemented on the text embeddings used in the image generation process. 
It first maps the text embedding of the attribute into a \textbf{unified latent space} (\textit{high-dimensional space}), where individual semantic components are represented by sparse neuron activations. By selectively activating these neurons, the module can reconstruct and control specific attributes in a disentangled manner through decoding. 
Specifically, we collect common textual prompts in portrait generation as our training dataset and encode them into the text embeddings.
Subsequently, the Attribute Sparse Autoencoder is trained to perform auto-encoding on these embeddings and activate as sparsely as possible, in which the sparsity is the key to attribute disentanglement.
Through this training, the unified latent space forms a compact, compositional dictionary of semantic components that can be flexibly recombined to represent new or unseen attributes without any retraining.
During inference, we can control the activation of corresponding components in the latent space. 
By directly manipulating the coefficient of activated components, it naturally supports continuous and fine-grained control over desired attributes.

Through operating in this sparsely activated latent space of attributes, our All-in-One Slider can achieve decoupled control of multiple attributes through a single module, without the need for any specific training for each attribute.
Benefiting from the compositional nature of the unified latent space, our All-in-One Slider generalizes to a wide range of attributes and enables continuous zero-shot manipulation of previously unseen ones.
Both qualitative and quantitative experiments are conducted to demonstrate the effectiveness of our All-in-One Slider for continuous and fine-grained manipulation of attributes.

The main contributions of this paper are as follows:
\begin{itemize}
    \item We propose a lightweight Attribute Sparse Autoencoder to achieve \textbf{precise and continuous manipulation} of individual or multiple attributes of image generation.
    \item We propose the All-in-One Slider framework, enabling unified manipulation covering a wide range of diverse attributes, \textbf{without requiring multiple training sessions}.
    \item Benefiting from constructing a unified latent space of diverse attributes, for the first time, our method shows the \textbf{zero-shot continuous control for unseen attributes}.
\end{itemize}

\section{Related Work}
\label{sec:related-work}

\noindent\textbf{Attribute Manipulation in T2I Diffusion Models.}
Achieving precise controllability has become a central goal in T2I diffusion models, enabling users to steer generation toward desired contents or styles. Early approaches, such as ControlNet~\cite{zhang2023adding,mou2024t2i}, introduce spatial guidance by conditioning on structural cues like edges, poses, or depth maps. These methods are effective at preserving geometric alignment but are inherently limited in controlling high-level semantic attributes, \textit{e.g.}, age, expression, or makeup, which are not explicitly represented in these inputs.

To achieve fine-grained semantic control beyond spatial guidance, researchers explored semantic manipulation in latent spaces (\textit{e.g.}, noise or $\mathcal{H}$-space). Methods like SEGA~\cite{brack2023sega} and Boundary Guidance~\cite{zhu2023boundary} enable editing without training, but lack fine-grained continuity or compositional control over semantic attributes. Later works shifted toward manipulating the text embedding space to achieve more direct semantic control. Prompt-based methods~\cite{hertz2022prompt,brooks2023instructpix2pix} intervene through text prompts and attention maps, but often suffer from discrete or coarse edits. 
Meanwhile, editing methods like Imagic~\cite{kawar2023imagic} and ITI-Gen~\cite{zhang2023iti} focus on aligning generation with user-provided images or preferences, but rely on image-text pairs and lack scalability for general attribute manipulation.
Recent approaches, like ConceptSlider~\cite{gandikota2024concept} and AttributeControl~\cite{baumann2025continuous} (AttControl), seek to enable continuous control by learning attribute vectors. However, for each attribute, they all require training a dedicated slider from lots of pairwise data, which leads to parameter redundancy accumulated by sliders and limits flexibility and scalability in practical application.  

To address these existing limitations, we propose a unified framework that discovers an interpretable and disentangled attribute latent space. This design enables smooth and scalable manipulation across multiple facial attributes without requiring retraining or paired labels, providing precise, continuous control in a plug-and-play manner.

\noindent\textbf{Sparse Autoencoders (SAEs).}
Sparse autoencoders~\cite{ng2011sparse,olshausen1997sparse,bengio2013representation} have been proven effective in learning interpretable representations in unsupervised settings~\cite{coates2011analysis}. Among them, K-Sparse Autoencoders (K-SAE)~\cite{makhzani2013k} enforce sparsity by allowing only the top-$k$ activations to activate, promoting the emergence of semantically meaningful units. 
Some studies in the context of large language models~\cite{gao2024scaling,huben2023sparse} introduce SAEs in capturing discrete semantic concepts.
Recent studies leverage the SAEs to interpret and analyze the internal mechanisms of diffusion models.
SAeUron~\cite{cywinski2025saeuron} uses SAEs for concept unlearning by suppressing undesired features across diffusion steps. Diffusion Lens~\cite{toker2024diffusion} decodes intermediate text encoder activations to reveal semantic representations. Unpacking SDXL Turbo~\cite{surkov2025unpacking} decomposes transformer blocks to expose interpretable, causal factors across style, structure, and semantics. However, these works primarily focus on model interpretability and causal tracing.
In this work, we extend K-SAE into a controllable face attribute manipulation framework that serves as a lightweight and scalable solution, enabling precise attribute control without retraining the base model.

\label{sec:Method}
\begin{figure*}[!h]
    \centering
    \includegraphics[width=1\linewidth]{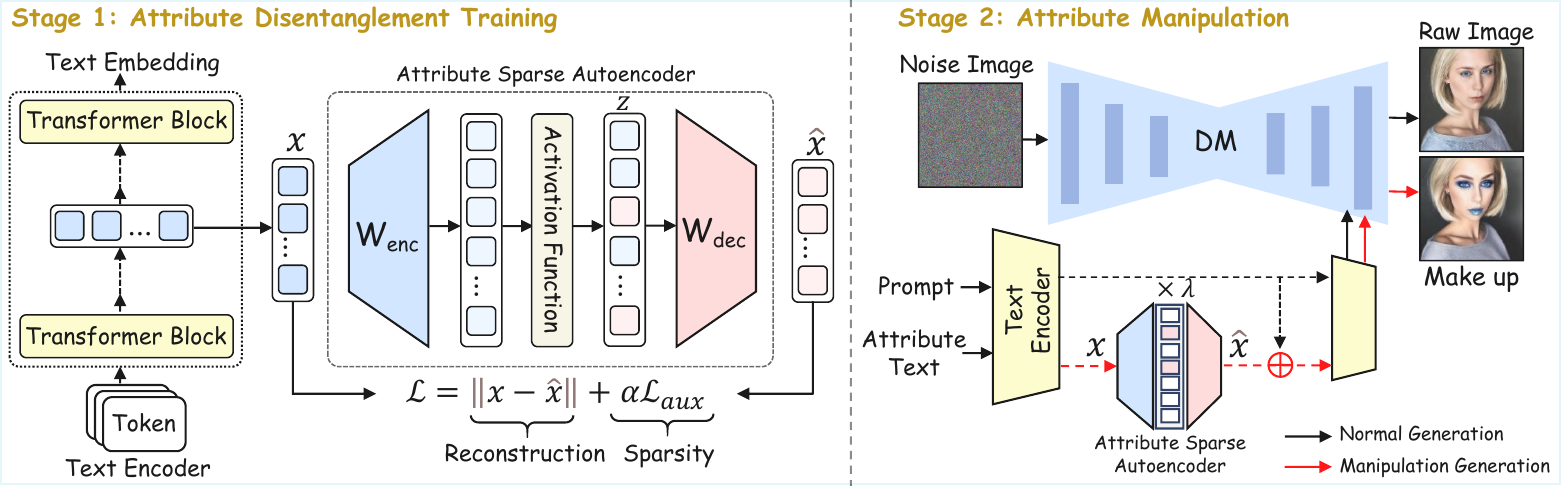} 
    \vspace{-8mm}
    \caption{An overview of our All-in-One Slider framework. 
    Stage 1: Unsupervised training of Attribute Sparse Autoencoder, which takes intermediate token embeddings from the residual stream in the text encoder as input and aims to reconstruct them with sparse features. Stage 2: Applying the trained Attribute Sparse Autoencoder
    to flexibly manipulate specific attributes during the image generation process.} 
    \vspace{-3mm}
    \label{fig:overview-structure} 
\end{figure*}
\section{Method}
To break away from the traditional One-for-One paradigm, where each attribute requires a dedicated module or control path, we propose the All-in-One Slider framework. The core architecture of our framework is a highly parameter-efficient module, the Attribute Sparse Autoencoder. It achieves attribute disentanglement modeling and constructs a unified attribute latent space. This design enables precise, composable, and generalizable control over multiple visual attributes without retraining or attribute-specific supervision.
As shown in Figure~\ref{fig:overview-structure}, the All-in-One Slider contains two key stages:  
(1) learning a latent representation from intermediate text embeddings, and  
(2) manipulating the diffusion generation process by modifying these latent codes. The full procedure is outlined in Algorithm~\ref{alg:k-SAE}.  

\subsection{Preliminary: Diffusion-based T2I Generation}
With the advent of DDPM~\cite{ho2020denoising}, diffusion models~\cite{saharia2022photorealistic,nichol2021glide} have become the dominant technology in visual generation, especially for many state-of-the-art T2I generation systems. Recent T2I diffusion models typically follow the latent generation process~\cite{rombach2022high} and consist of three main components: a Variational Auto-Encoder (VAE)~\cite{kingma2013auto}, a text encoder, and a diffusion generator. The VAE compresses images into a latent space and decodes the generated latent back to pixel space. The text encoder, such as CLIP~\cite{radford2021learning}, encodes textual prompts into embeddings. The diffusion generator then takes a noisy latent as input and captures prompt information through cross-attention with the text embeddings, iteratively denoising the noisy latent towards a final output, enabling text-to-image generation.

\begin{algorithm}[!t]
\caption{Train Slider \& Attribute Manipulation}
\label{alg:k-SAE} 
\begin{algorithmic}[1]
\renewcommand{\algorithmicrequire}{\textbf{Input:}}
\renewcommand{\algorithmicensure}{\textbf{Output:}}

\REQUIRE Text prompt $x_{\text{txt}}$, attribute text $A$, scalar $\lambda$
\ENSURE Manipulated image $I$
\vspace{0.4em}

\STATE Initialize ENC and DEC  
\STATE $\mathcal{D} \leftarrow \mathcal{E}(x_{\text{txt}})$  

\vspace{0.4em}
\item[] \hspace{-1.2em} \textit{Stage 1: Attribute Disentanglement Training}
\FOR{$i \leftarrow 1$ to $N$}
    \STATE $x \leftarrow \mathcal{D}$
    \STATE $z_{\text{ALS}} \leftarrow \text{Top-}k(\operatorname{ReLU}(W_{\text{enc}}(x - b_{\text{pre}}) + b_{\text{enc}}))$
    \STATE $\hat{x} \leftarrow W_{\text{dec}} z_{\text{ALS}} + b_{\text{pre}}$
    \STATE $\mathcal{L} \leftarrow \| x - \hat{x} \|^2_2 + \alpha \mathcal{L}_{\text{aux}}$ 
\ENDFOR

\vspace{0.4em}
\item[] \hspace{-1.2em} \textit{Stage 2: Attribute Manipulation}
\STATE $x_{\text{A}} \leftarrow \mathcal{E}(A)$
\STATE $z^{\text{A}}_{\text{ALS}} \leftarrow \operatorname{ENC}(x_{\text{A}})$
\STATE $x_{\text{manipulated}} \leftarrow x + W_{\text{dec}}(\lambda \times z^{\text{A}}_{\text{ALS}})$
\STATE $I \leftarrow \operatorname{DM}(x_{\text{manipulated}})$ 
\STATE \textbf{return} $I$
\end{algorithmic}
\end{algorithm}


\subsection{Attribute Disentanglement Training}
\label{subsec:saetrain}
To achieve encoding and memorization of as many attributes as possible with limited parameters, we introduce the Attribute Sparse Autoencoder, which contains an encoder and a decoder.
For training, we collect numerous text embeddings from the text encoder of a pretrained diffusion model.
The encoder part of the Attribute Sparse Autoencoder maps these embeddings into a high-dimensional latent space (referred to as \textit{Attlatentspace} in the following).
Once trained, this \textit{Attlatentspace} can serve as a unified attribute space, where attributes with different semantics sparsely activate a minimal and distinct set of neurons, thereby enabling disentangled modeling of diverse attributes.
The decoder part then reconstructs the original embeddings from the latent representations. To encourage sparsity and disentanglement in the \textit{Attlatentspace},
we draw inspiration from SAEs~\cite{makhzani2013k, gao2024scaling, karvonen2024measuring} and apply a top-$k$ activation strategy to sparsely activate the attribute latent space.
The overall training procedure consists of three major steps: embedding extraction, sparse encoding, and sparse reconstruction, as detailed below.

\noindent\textbf{Embedding Extraction.} The text encoder $\mathcal{E}$ of a pretrained diffusion model DM (\textit{e.g.}, SDXL) consists of a stack of transformer blocks $\{B_1, \ldots, B_L\}$. We extract the hidden state $x$ from an intermediate block $B_l$. Specifically, we extract features from the dual text encoders of SDXL, concatenate them to form the final joint representation $x \in \mathcal{D}$, and feed it into the Attribute Sparse Autoencoder.

\noindent\textbf{Sparse Encoding.} 
We implement sparsity using a top-$k$ selection mechanism. The encoder $\text{ENC}$ maps the extracted  embedding $x$ into the \textit{Attlatentspace} $z_{\text{ALS}}$ via a linear transformation, ReLU activation, and top-$k$ selection:
\begin{equation}
z_{\text{ALS}} = \text{Top-}k\left(\text{ReLU}(W_{\text{enc}}(x - b_{\text{pre}}) + b_{\text{enc}})\right).
\label{eq:encoder_step1}
\end{equation} 
Here, $b_{\text{pre}}$ and $b_{\text{enc}}$ denote learnable bias terms. Retaining only the top-$k$ neurons with the highest activations yields a compact and disentangled representation, where each attribute is associated with a distinct subset of neurons. The sparsity is applied to the individual channels of the text embedding, enabling the selective activation of features that are most pertinent to the semantics of specific attributes in the high-dimensional embedding space (\textit{i.e., Attlatentspace}). This channel-wise sparsity thereby promotes attribute disentanglement, allowing the model to isolate fine-grained semantic signals while suppressing irrelevant noise.
The decoder $\text{DEC}$ then reconstructs $\hat{x}$ as:
\begin{equation}
\hat{x} = W_{\text{dec}} z_{\text{ALS}} + b_{\text{pre}}.
\label{eq:decoder}
\end{equation}

\noindent\textbf{Sparse Reconstruction.}
We train Attribute Sparse Autoencoder to minimize the reconstruction loss:
\begin{equation}
\mathcal{L}_{\text{mse}} = \| x - \hat{x} \|^2_2.
\label{eq:loss}
\end{equation}
However, a known limitation of sparse autoencoders is the prevalence of \textit{dead neurons}—latent units that often remain inactive throughout training. These unused neurons reduce the expressiveness of the learned space and hinder future composability.
To address this, we adopt an auxiliary mechanism inspired by~\cite{gao2024scaling}. At each training step, 
we compute the residual $r = x - \hat{x}$ and identify a set of $k_{\text{aux}}$ least-activated neurons. The corresponding activations, denoted as $\hat{z}_{\text{ALS}}$, are used to reconstruct the residual:
\begin{equation}
\hat{z}_{\text{ALS}} = \text{Top-}k_{\text{aux}}\left(\text{ReLU}(W_{\text{enc}}(x - b_{\text{pre}}) + b_{\text{enc}})\right).
\end{equation}
We compute the auxiliary reconstruction $\hat{r}$ as $\hat{r} = W_{\text{dec}} \hat{z}_{\text{ALS}}$, and define the auxiliary loss as:
$\mathcal{L}_{\text{aux}} = \| r - \hat{r} \|^2_2.$
The final training objective thus becomes:
\begin{equation}
\mathcal{L} = \mathcal{L}_{\text{mse}} + \alpha \mathcal{L}_{\text{aux}}.
\label{eq:L_sae}
\end{equation}
This loss encourages even initially inactive neurons to explain parts of the residual, ensuring broader semantic coverage in the latent space.
After training, latent units in the sparse latent space correspond to specific, disentangled facial attributes. This latent space serves as the basis for attribute manipulation, as detailed in the next section.

\subsection{Attribute Manipulation}
Once the Attribute Sparse Autoencoder is trained, it can be employed as an ``Attribute Manipulation Slider'', enabling control over specific attributes within the unified Attribute Latent Space of the text encoder. Given an attribute $A$ that we expect to manipulate, we extract its corresponding text embedding $x_\text{A}$ and feed it to the trained Attribute Sparse Autoencoder to identify the latent direction associated with the attribute. We then modify the input prompt embedding $x$ by adding this direction scaled by a scalar $\lambda$, as follows:
\begin{equation}
x_{\text{manipulated}} = x + W_{\text{dec}}(\lambda \times \text{ENC}(x_\text{A})),
\end{equation}
where $ \lambda$ controls the strength of the steering. This modification of the embedding $ x_{\text{manipulated}}$ is then used to condition the generation process, enabling precise manipulation of specific attributes without disrupting the overall quality or structure of the generated image.
Importantly, this manipulation operates entirely within the learned sparse latent space, supporting zero-shot, composable editing across a wide variety of attributes. This unified mechanism underpins the scalable and flexible attribute control explored. 

\section{Experiments}
\label{sec:experiments}
\subsection{Experimental Setup}
\noindent\textbf{Implementation Details.} We employ SDXL~\cite{podell2023sdxl} to generate images with 50 sampling steps with classifier-free guidance~\cite{ho2022classifier} of 7.5. All compared methods are implemented following their default configurations from their official repository. Besides, we edit real images using SDXL-Turbo~\cite{sauer2024adversarial} with the Renoise inversion method~\cite{garibi2024renoise}. We extract features from the 11th and 29th layers of the two text encoders in SDXL, which contain 12 and 32 transformer layers, respectively. We train our Attribute Sparse Autoencoder with 52,000 text samples (52 controllable facial attributes, and 1000 samples per attribute). The details of construction and Attribute Sparse Autoencoder parameters used for the experiments are included in the appendix. 


\noindent\textbf{Quantitative Metrics.} 
Traditional metrics, such as the CLIP score~\cite{hessel2021clipscore}, are limited in their ability to evaluate fine-grained or localized attribute manipulations. These methods typically measure overall semantic similarity, which may obscure whether specific target attributes were correctly edited.
To more comprehensively assess image editing quality and manipulation capabilities, we adopt two complementary metrics specifically designed to capture the nuanced effects of facial attribute manipulation. We use the multimodal large language model, Qwen2.5-VL-Instruct~\cite{Qwen2-VL}, to evaluate the semantic alignment between the intended textual prompt (\textit{e.g.}, ``smile'', ``old'', and ``makeup'') and the generated image, termed as \textbf{Qwen Score (QS)}~\cite{han2025ntire}. In contrast to traditional metrics, Qwen demonstrates superior reasoning ability and aligns more closely with human judgment. The QS evaluates the edited image relative to the original image based on three criteria: Action Fidelity (the alignment with the edit prompt), Identity Preservation (the consistency of the subject’s identity), and Visual \& Anatomical Coherence (the naturalness of the modifications). The average of these three scores is used as the final metric. Moreover, we further measure the identity preservation between the source image and the steered result using an \textbf{ID Consistency Score (IS)} based on a face recognition model (\textit{i.e.}, Arcface~\cite{deng2019arcface}). This quantifies how well our method preserves the core identity features despite attribute manipulations. Identity preservation is especially important in real-image editing scenarios, where users expect natural-looking changes without sacrificing personal identity. Higher QS and IS both indicate better editing performance, reflecting more accurate attribute manipulation and stronger identity preservation.

\begin{figure}[t]
    \centering
    \includegraphics[width=0.9\linewidth]{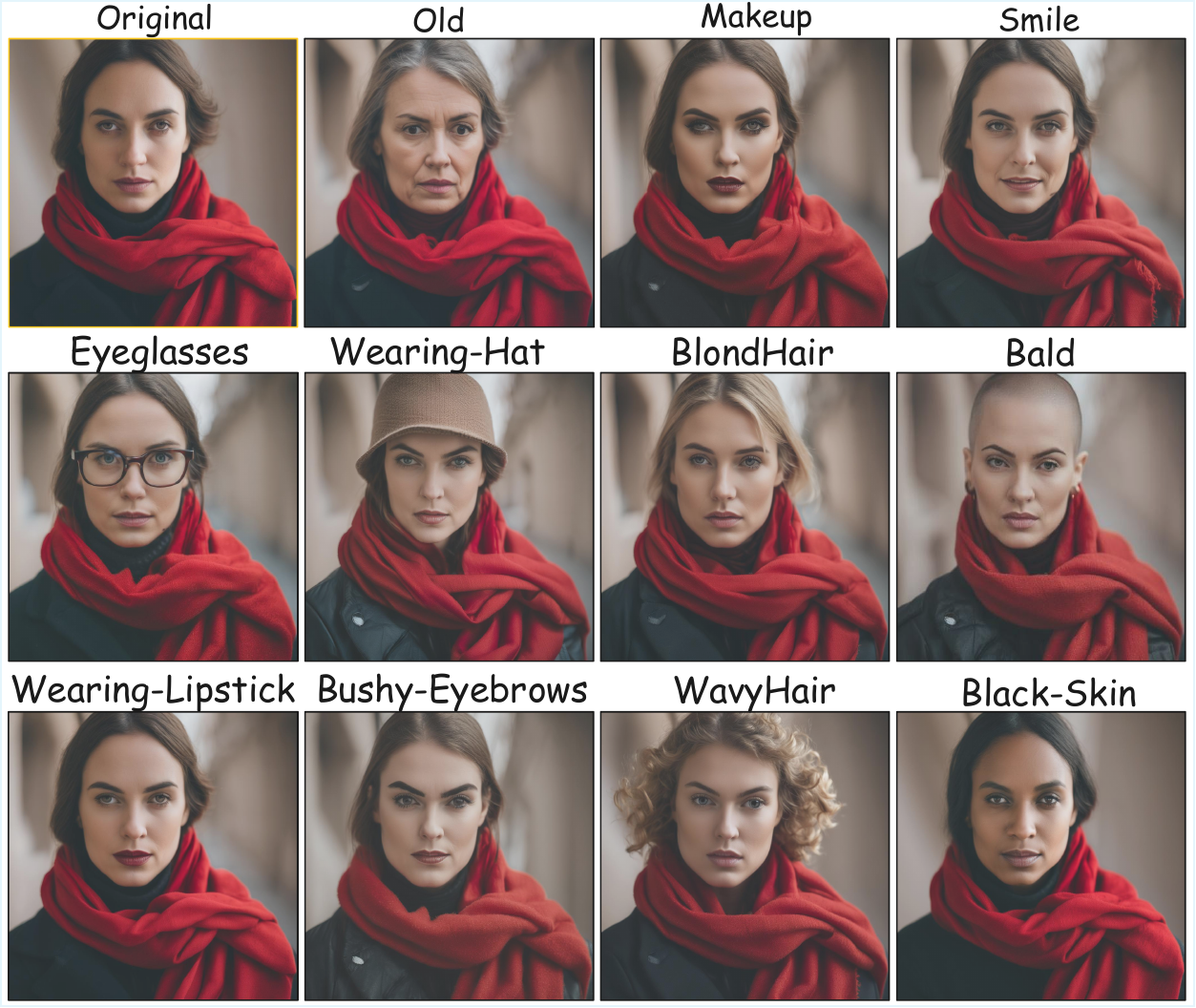} 
    \vspace{-4mm}
    \caption{Qualitative results of face attribute manipulation. Our All-in-One slider can perform both fine-grained semantic edits (\textit{e.g.}, smile, makeup, and age) and physical changes (\textit{e.g.}, eyeglasses, hat, hair style, and skin tone).} 
    \vspace{-2mm}
    \label{fig:qualitative-result-fineAtt} 
\end{figure}
\subsection{Text-to-Image Attribute Manipulation}
To evaluate the ability of All-in-One Slider to manipulate attributes, we present qualitative results in Figure~\ref{fig:qualitative-result-fineAtt}. The wide range of visual controls, which covers both semantic attributes (\textit{e.g.}, \textit{smile}, \textit{age}, \textit{makeup}) and physical accessories (\textit{e.g.}, \textit{eyeglasses}, \textit{hat}), demonstrates the expressive power and controllability of our method. Each manipulated image preserves the identity and background of the original input, while exhibiting precise and consistent attribute changes. 
This controllability is enabled by the sparsity and disentanglement of directions in the latent space of Attribute Sparse Autoencoder, where each attribute corresponds to an independent and well-isolated latent direction. This allows precise manipulation without affecting unrelated visual features. In contrast, existing methods~\cite{baumann2025continuous,gandikota2024concept} rely on individual sliders per attribute, which inherently restricts the controllable space as each new attribute requires its own slider. 
Our unified framework overcomes this constraint, enabling efficient manipulation across broader attributes.

\begin{figure}[t]
    \centering
    \includegraphics[width=1.0\linewidth]{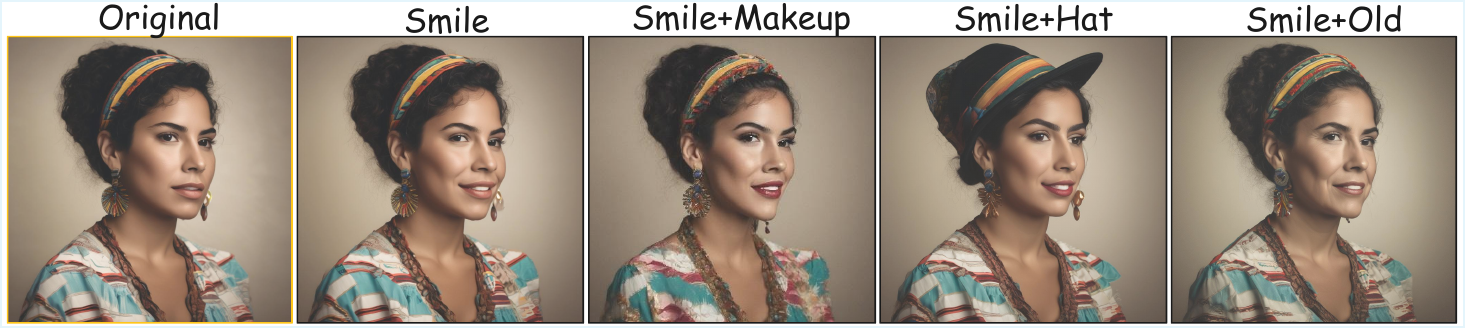} 
    \vspace{-7mm}
    \caption{Qualitative results of compositional multi-attributes manipulation. Our All-in-One slider achieves coherent manipulation while preserving the original identity.}
    \vspace{-3mm}
    \label{fig:attAdd} 
\end{figure}
\noindent\textbf{Multi-Attribute Manipulation.}
To evaluate the compositional controllability of our method, we perform multi-attribute manipulation experiments as shown in Figure~\ref{fig:attAdd}. Starting from a base image, we simultaneously apply multiple attributes such as \textit{Smile}, \textit{Makeup}, \textit{Hat}, and \textit{Old}.
The edited images show a visually coherent and semantically consistent blend of attributes without artifacts or conflicts, demonstrating the model’s ability to naturally combine multiple attributes while still preserving the original identity.
Similar compositional behavior is also observed in Figure~\ref{fig:intro-visuli} (2), where various attribute combinations (\textit{e.g.}, \textit{Smile}, \textit{Old}, \textit{Eyeglasses}, \textit{WavyHair}) are applied, yielding realistic and conflict-free results across identities. 
This compositional behavior stems from our unified latent space of Attribute Sparse Autoencoder, where different attribute directions are effectively disentangled yet compatible. Since all manipulations are represented as sparse vectors within a shared embedding space, their combinations remain semantically coherent and do not interfere with each other.
\begin{table}[tbp]
\caption{Quantitative comparisons of single and multiple attributes editing.
``QS'' denotes the semantic alignment score by Qwen, and ``IS'' denotes the identity consistency score.
The best and second-best results are highlighted with \textbf{bold} and \underline{underline}, respectively.
}
\vspace{-3mm}
\small
\centering
\setlength{\tabcolsep}{2.7pt} 
\begin{tabular}{llcccccc}
\toprule
\multirow{2}{*}{Set} & \multirow{2}{*}{Method} & \multicolumn{2}{c}{Old} & \multicolumn{2}{c}{Smile} & \multicolumn{2}{c}{Makeup} \\
\cmidrule(lr){3-4} \cmidrule(lr){5-6} \cmidrule(lr){7-8}
& & QS & IS & QS & IS & QS & IS \\
\midrule
\multirow{3}{*}{Sng} 
  & CSlider & 3.7941 & 0.4336 & 4.1437 & 0.4962 & \textbf{4.5424} & \underline{0.6529} \\
  & AttCtrl & \underline{4.0392} & \underline{0.6005} & \textbf{4.3954} & \textbf{0.6952} & 4.2680 & 0.6044 \\
  & Ours    & \textbf{4.0490} & \textbf{0.7155} & \underline{4.2647} & \underline{0.6366} & \underline{4.2908} & \textbf{0.7423} \\
\midrule
\multicolumn{2}{c}{} & \multicolumn{2}{c}{Old+Smile} & \multicolumn{2}{c}{Old+Makeup} & \multicolumn{2}{c}{Smile+Makeup} \\
\cmidrule(lr){3-4} \cmidrule(lr){5-6} \cmidrule(lr){7-8}
\multirow{3}{*}{Mul} 
  & CSlider & \underline{4.1503} & \underline{0.4993} & 3.8006 & 0.5224 & 4.0588 & 0.4786 \\
  & AttCtrl & 3.6667 & 0.3755 & \underline{4.0555} & \textbf{0.6347} & \underline{4.2484} & \underline{0.5145} \\
  & Ours    & \textbf{4.2124} & \textbf{0.6882} & \textbf{4.4281} & \underline{0.6277} & \textbf{4.2973} & \textbf{0.6351} \\
\bottomrule
\end{tabular}
\vspace{-5mm}
\label{tab:base-results}
\end{table}

\noindent\textbf{Quantitative Analysis.}
We quantitatively evaluate the effectiveness of our method in both single-attribute (Sng) and multi-attribute (Mul) manipulating tasks, using QS
and IS as evaluation metrics. We compare against two One-for-One baselines: ConceptSlider~\cite{gandikota2024concept} and AttributeControl~\cite{baumann2025continuous}. Results are summarized in Table~\ref{tab:base-results}.
In the single-attribute setting, our method achieves competitive performance, obtaining the highest IS for both \textit{Old} and \textit{Makeup} attributes, and ranking first in QS for the Old attribute and second for Smile and Makeup.
In contrast, in the multi-attribute setting—where compositional complexity and entanglement pose greater challenges—our method outperforms the baselines. It achieves the highest QS across all combinations (\textit{Old+Smile}, \textit{Old+Makeup}, and \textit{Smile+Makeup}) and the highest IS scores in two out of three cases. Notably, for the challenging \textit{Old+Makeup} case, our approach achieves a QS of 4.4281, surpassing the second-best baseline (4.0555) by a clear margin while still achieving a highly competitive IS of 0.6277. 

These results highlight the core strength of our approach: the ability to perform disentangled, compositional steering while preserving identity. Unlike prior methods that often sacrifice attribute fidelity for identity preservation, our method strikes a better balance, maintaining both attribute accuracy and subject identity in both single- and multi-attribute manipulation tasks. The superior performance, particularly in multi-attribute cases, stems from our disentangled Attribute Latent Space, which enables conflict-free, compositional control over multiple attributes without compromising the subject’s core identity.

\begin{figure}[t]
    \centering
    \includegraphics[width=0.9\linewidth]{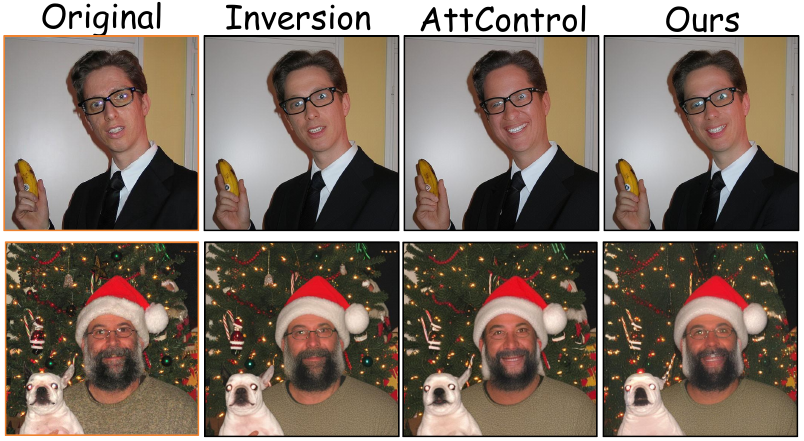} 
    \vspace{-3mm}
    \caption{Real image editing. We apply our method to edit Smile (top) and Old (bottom) attributes using the ReNoise inversion framework~\cite{garibi2024renoise}. Compared to AttControl, our method better preserves identity details like eyeglasses and facial structure.} 
    \vspace{-5mm}
    \label{fig:sdxl_real_imgedit} 
\end{figure}
\subsection{Reference-Based Image Editing}
To enable high-fidelity editing on real images, we integrate our method with the ReNoise inversion framework~\cite{garibi2024renoise}, which reconstructs latent representations from real inputs. As illustrated in Figure~\ref{fig:sdxl_real_imgedit}, we conduct editing on two representative attributes: \textit{Smile} (top) and \textit{Old} (bottom). 
In the first row, when adding the \textit{Smile}, our method clearly avoids unnatural facial deformation (\textit{e.g.}, overly rounded cheeks observed in AttControl). In the second case, it successfully preserves accessories like eyeglasses during the addition of complex attributes, such as combining Christmas elements with the \textit{old} attribute.
Compared to AttributeControl, our approach better maintains critical identity features. This is primarily attributed to our Attribute Sparse Autoencoder-based decoupling strategy, which minimizes unintended attribute entanglement and improves the overall stability and locality of edits via sparse semantic modulation.


\begin{figure}[!tb]
    \centering
    \includegraphics[width=0.9\linewidth]{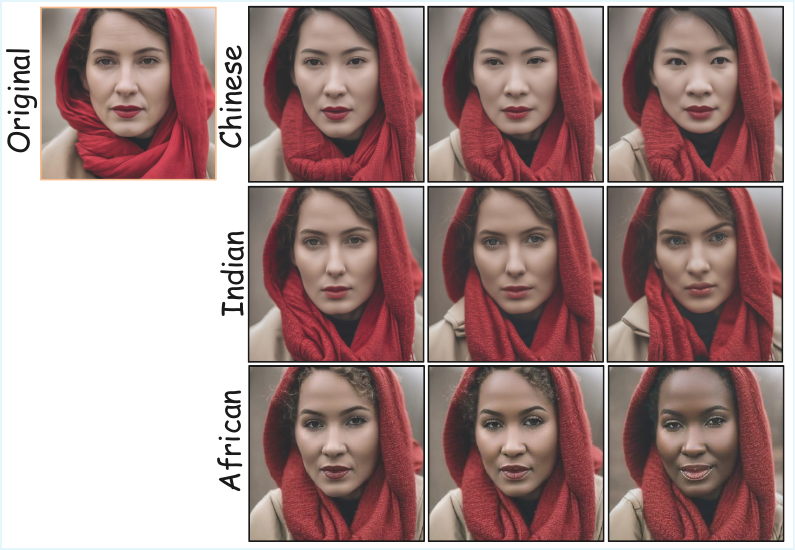} 
    \vspace{-4mm}
    \caption{Continuous zero-shot generalization to different racial attributes. The model progressively manipulates facial features across varying strengths of the racial attribute.}
    \vspace{-5.5mm}
    \label{fig:zero-shot-race} 
\end{figure}

\subsection{Zero-shot Generalization Ability}
Although our All-in-One Slider is trained on a fixed set of facial attributes, it demonstrates strong zero-shot generalization to new racial and identity attributes not encountered during training. As shown in Figure~\ref{fig:zero-shot-race}, the model can continuously modify facial attributes, including races (\textit{e.g.}, African, Chinese and Indian) and celebrity-like identities (\textit{e.g.}, Barack Obama and Einstein in Figure~\ref{fig:intro-visuli} (3)), without the need for additional fine-tuning or supervision. 

This generalization ability arises from our attribute decomposition strategy, where each facial attribute is learned as a set of fine-grained, disentangled components within the shared \textit{Attlatentspace}.
The pretrained text encoder has built the alignment of generalized objects and their attribute correspondences.
For objects  (\textit{e.g.}, races and celebrity names) that have not been seen during training but are memorized by the generative model itself, their text embedding can still induce adaptive combinatorial activation of attribute components within \textit{Attlatentspace} to reconstruct corresponding complex combinations of key attributes.
This ability to perform zero-shot and open-vocabulary manipulation of attributes further underscores the versatility of our approach for controllable image synthesis.

\subsection{Applications}
\noindent\textbf{Model Generalization.}
We further investigate the generalization of our method by applying it to various diffusion models, including Stable Diffusion v1.4 (SD v1.4), SDXL-Turbo and the Diffusion Transformer (DiT) (see Appendix). In Figure~\ref {fig:sd1.4_makeup} (1), progressively increasing the control strength $\lambda$ yields fine-grained variations in \textit{makeup} intensity, demonstrating effective attribute modulation within the SD v1.4. Figure~\ref{fig:sd1.4_makeup} (2) further validates our method on SDXL-Turbo, where semantic manipulation of the \textit{Smile} attribute is achieved with high fidelity and robust identity preservation. These results highlight the robustness and transferability of our approach across diverse backbones.
\begin{figure}[t]
    \centering
    \includegraphics[width=0.9\linewidth]{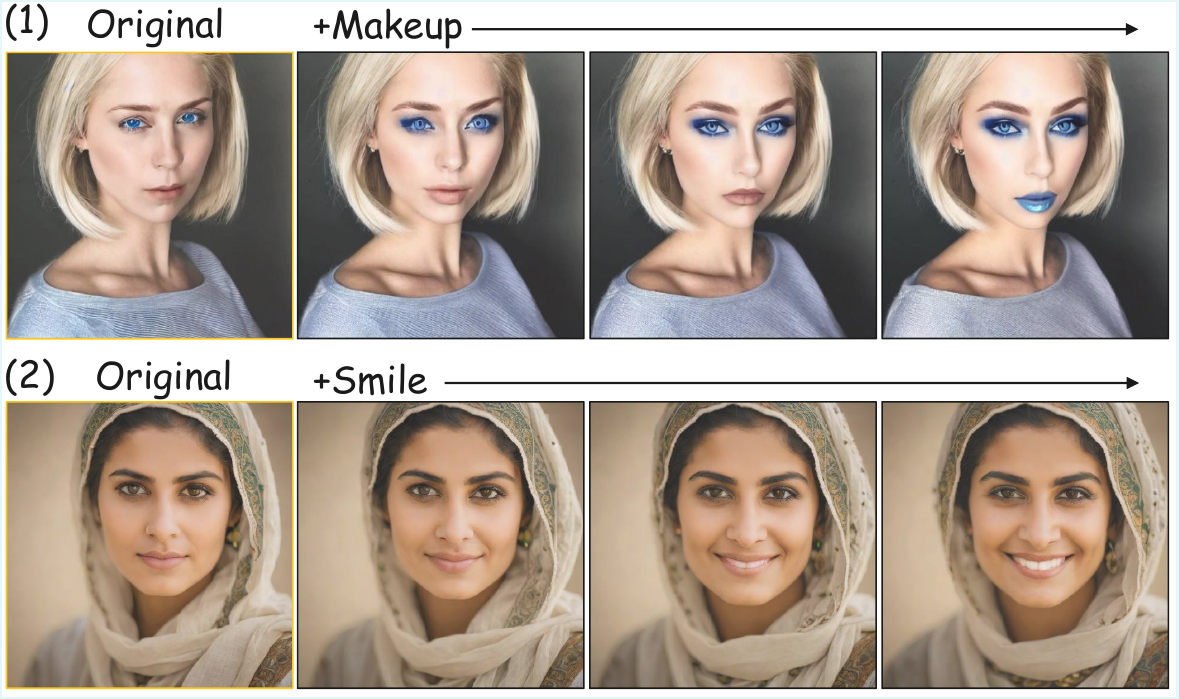} 
    \vspace{-4mm}
    \caption{Qualitative results on SD v1.4 (1) and SDXL-Turbo (2) showing the effectiveness of our method in attribute modulation (\textit{e.g.}, makeup and smile) across U-Net-based architectures.} 
    \vspace{-5mm}
    \label{fig:sd1.4_makeup} 
\end{figure}

\noindent\textbf{Style Manipulation.}
To validate our method's versatility beyond facial attributes manipulation, we apply it to photography style steering. We train SAE on 40 different photography styles following the procedure in Sec.~\ref{subsec:saetrain}, including \textit{Black-and-White}, \textit{Golden-hour}, and \textit{Fisheye}, etc. As shown in Figure~\ref{fig:app-style}, our method effectively transforms the original image into various photography styles while preserving the core content and composition. 

\begin{figure}[t]
    \centering
    \includegraphics[width=1.0\linewidth]{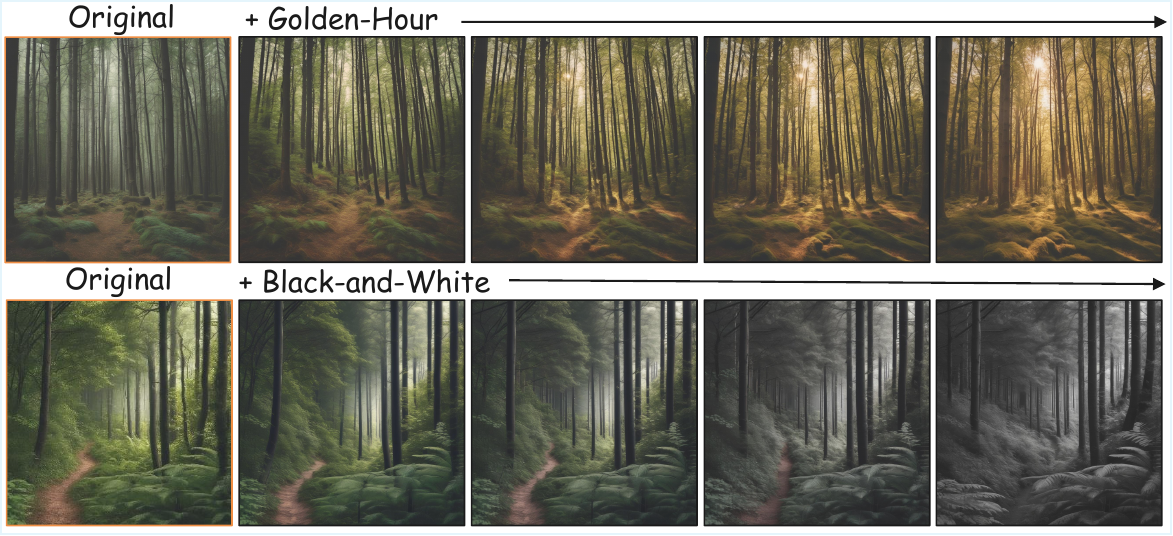} 
    \vspace{-8mm}
    \caption{Style manipulation on SDXL: the original image is progressively transformed into \textit{Black-and-White} and \textit{Golden-hour} styles while preserving main content and composition.} 
    \vspace{-6mm}
    \label{fig:app-style} 
\end{figure}

\noindent\textbf{Multi-Subject Attribute Manipulation.} 
To extend the universal Attribute Sparse Autoencoder (detailed in Sec.~\ref{subsec:saetrain}) to multi-subject scenarios, we introduce an \textit{Attention Pooling (AttPooling) Aggregator} (AAg) module. It aggregates the latent features of attribute semantics ($z^\pm$) onto a token, enabling precise localization of the target subject (\textit{e.g.}, ``woman'' or ``man'').
To extract pure attribute directions in this setting, we create 4,000 sentence pairs, with each pair consisting of one sentence describing a specific attribute (\textit{e.g.}, \textit{smile}, \textit{old}, \textit{makeup}) and another without it. 
Building upon the pretrained universal SAE (\textit{i.e.}, trained in \textit{stage 1} in Sec.~\ref{subsec:saetrain}), we jointly fine-tune the SAE and the AAg module with SDXL using the following objective: $\mathcal{L}_{\text{multi}} {=} \mathcal{L}_{\text{sae}} + \eta \mathcal{L}_{\text{cons}}$. Here, $\mathcal{L}_{\text{sae}}$ denotes the original sparse and reconstruction loss (Eq.~\ref{eq:L_sae}), while $\mathcal{L}_{\text{cons}}$ is a consistency loss that encourages the model to preserve non-target regions. This fine-tuning process enhances the model’s ability to apply attributes selectively to specified subjects. The overall fine-tuning framework for multi-subject manipulation is illustrated in Figure~\ref{fig:multi-sub}(a) (detailed in Appendix). As shown in Figure~\ref{fig:multi-sub}(b), our method can effectively manipulate attributes for specific subjects (\textit{e.g.}, adding \textit{old} to the man, and \textit{smile} to the woman) while maintaining the identity and context of other elements in the scene.



\begin{figure}[t]
    \centering
    \includegraphics[width=0.9\linewidth]{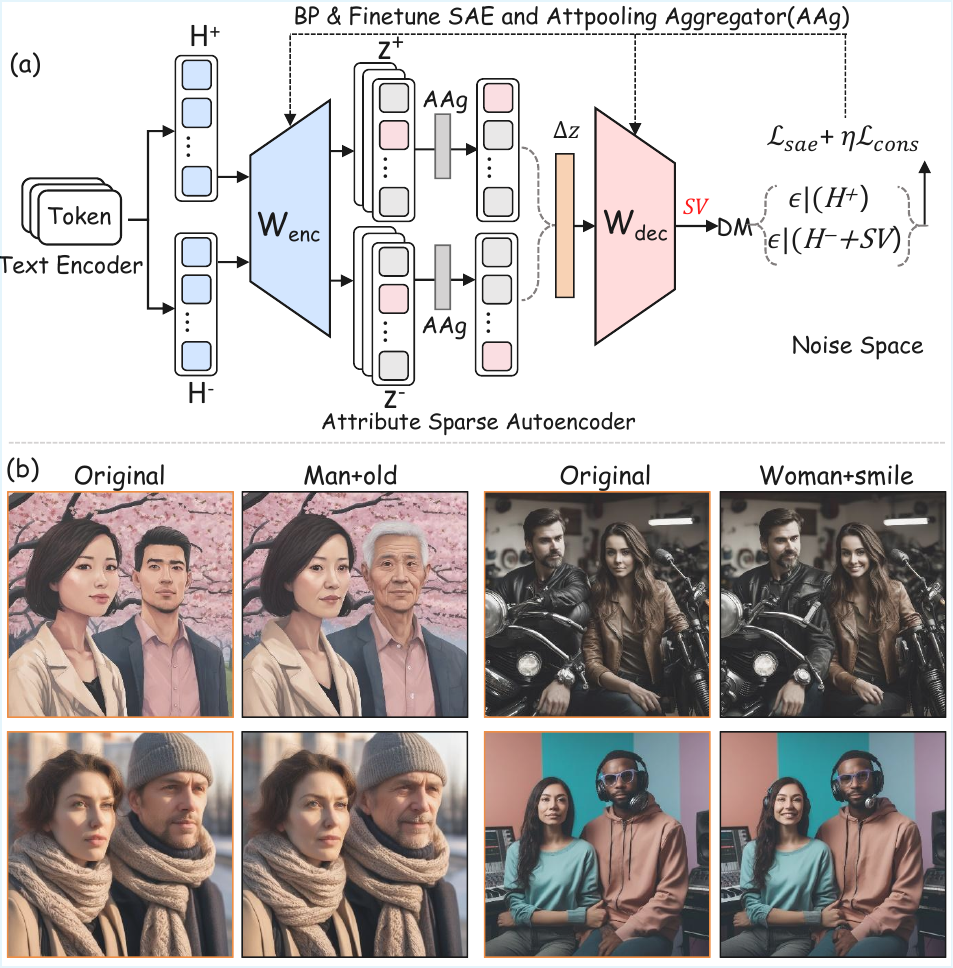} 
    \vspace{-4mm}
    \caption{(a) Framework for fine-tuning multi-subject manipulation. The Attribute Sparse Autoencoder is fine-tuned using paired sentences combined the diffusion model, and we introduce an \textit{Attpooling Aggregator} (AAg) module to locate the target subject for manipulation. (b) Qualitative results of applying attributes to the targeted subject (\textit{e.g.}, \textit{old} to the man, \textit{smile} to the woman).} 
    \vspace{-5mm}
    \label{fig:multi-sub} 
\end{figure}




\subsection{Ablation Study}
We conduct ablation studies on two key components of our method: the manipulated text encoder layers and the steering strength used during attribute control.

\noindent\textbf{Layer Selection.}  
We evaluate performance across various layer configurations of SDXL's dual text encoders, specifically testing different combinations of layer indices (X/Y) as shown in Table~\ref{tab:layer-select}. These layers exert a pronounced impact on the generated images, capturing different levels of semantic representation. The results indicate that deeper layers generally yield higher QS (semantic score), while IS tend to decline. The 10/28 combination achieves the best overall balance, indicating that mid-level layers are more effective at simultaneously preserving semantic intent and identity. This clearly reflects the practical importance of choosing appropriate feature granularity for localized and identity-sensitive editing in practice.

\begin{figure}[t]
    \centering
    \includegraphics[width=0.95\linewidth]{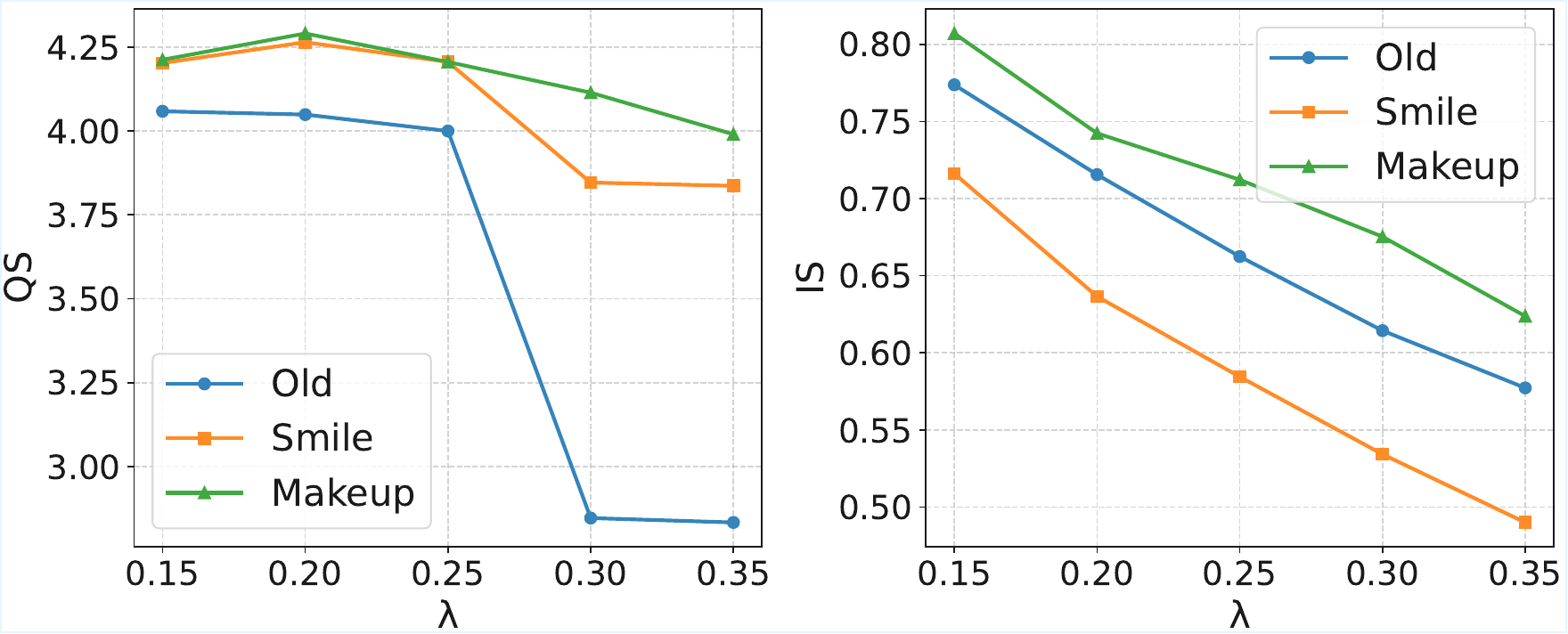} 
    \vspace{-4mm}
    \caption{Effect of manipulating strength $\lambda$ on attribute expression (QS) and identity preservation (IS) for various attributes.}
    \vspace{-5mm}
    \label{fig:lambda_ablation} 
\end{figure}

\noindent\textbf{Effect of Manipulating Strength $\lambda$.}  
Figure~\ref{fig:lambda_ablation} illustrates how varying $\lambda$ affects the trade-off between attribute control and identity preservation. Lower values (\textit{e.g.}, $\lambda=0.15$) lead to under-editing, whereas higher values (\textit{e.g.}, $\lambda=0.30$) result in stronger attribute expression. 
In particular, semantic degradation is most pronounced for the \textit{old}, suggesting that aging transformations are more deeply entangled with identity features. 
These results demonstrate the inherent trade-off between attribute modification and identity preservation during attribute manipulation.


\begin{table}[t]
\caption{Evaluation for attributes across different Manipulation Layer (X/Y) of SDXL's dual text encoders, where X is the selected layer index in the first encoder and Y in the second.}
\vspace{-3mm}
\small
\centering
\setlength{\tabcolsep}{2pt} 
\begin{tabular}{c|cc|cc|cc|cc}
\toprule
\multirow{2}{*}{X/Y} & \multicolumn{2}{c|}{\textbf{Old}} & \multicolumn{2}{c|}{\textbf{Smile}} & \multicolumn{2}{c|}{\textbf{Makeup}} & \multicolumn{2}{c}{\textbf{Avg}} \\
& QS & IS & QS & IS & QS & IS & QS & IS \\
\midrule
8/28  & 3.961 & 0.672 & 4.216 & 0.537 & 4.196 & 0.696 & 4.124 & 0.635 \\
9/30  & 4.062 & 0.631 & 4.193 & 0.617 & 4.177 & 0.759 & 4.144 & 0.669 \\
10/24 & 4.023 & 0.722 & 4.245 & 0.668 & 4.288 & 0.765 & \underline{4.185} & \underline{0.718} \\
10/28 & 4.049 & 0.716 & 4.265 & 0.637 & 4.291 & 0.742 & \textbf{4.202} & \textbf{0.698} \\
\bottomrule
\end{tabular}
\vspace{-3mm}
\label{tab:layer-select}
\end{table}

\section{Conclusion}
\label{sec:discussion and conclusion}
We propose the All-in-One Slider, a sparse and lightweight framework for controllable attribute manipulation using the Attribute Sparse Autoencoder.
Unlike prior approaches that rely on attribute-specific training, our method introduces a unified and disentangled latent space, enabling fine-grained, continuous manipulation of multiple facial attributes while preserving identity and visual fidelity. 
Extensive experiments demonstrate performance of our method in both single- and compositional-attribute manipulation, as well as zero-shot generalization to unseen attributes. All-in-One Slider presents a new and unified paradigm for continuous attribute control of image generation, offering scalable, interpretable, and flexible advantages. This motivates further work on continuous attribute control with low training and parameter costs based on the unified space. 

\noindent \textbf{Acknowledgments.} This work has been supported by the Fundamental Research Funds for the Central Universities (No. 2022XKRC015), Natural Science Foundation of Beijing Municipality (No. L252025), National Natural Science Foundation of China (No. 62372033), and the Science and Technology Development Fund of Macao SAR under the International Collaborative Research (No. 0001/2025/AIJ).

{
    \small
    \bibliographystyle{ieeenat_fullname}
    \bibliography{main}
}

\newpage

\appendix

\section{Implementation Details}
\paragraph{Training Details.}
We use SDXL as the base generative model and extract intermediate text embeddings from its 11th (of 12) and 29th (of 32) transformer layers blocks (\texttt{layer.10} and \texttt{layer.28}) across the dual encoders to train the Attribute Sparse Autoencoder. The activations are computed over prompts covering 52 attributes, each augmented with diverse identity-aware prefixes to encourage broad semantic coverage.
The dimension of text embedding is 2048 (concatenation of 768 and 1280). The autoencoder has a latent dimension of 32,768 with an expansion factor of 16. It is trained using the Adam optimizer with a learning rate of $4 \times 10^{-4}$ and a batch size of 4096 latent tokens per step. The total training budget is 400 million tokens, corresponding to approximately 97,656 training steps. 
During attribute selection, we retain the top-$k=128$ most activated dimensions. To mitigate overfitting and encourage broader attribute coverage, we introduce an auxiliary selection mechanism with $k_{\text{aux}} = 256$ and apply a regularization coefficient of $\alpha = 0.1$ to increase its influence during training. All experiments were performed using a single NVIDIA GeForce RTX 4090 GPU.

\paragraph{Training Prompt Construction.}
To expose the Attribute Sparse Autoencoder to a wide variety of visual concepts, we construct a diverse set of text prompts associated with over 52 semantic attributes (\textit{e.g.}, \textit{age}, \textit{arched eyebrows}, \textit{wearing necklace}). For each attribute, we manually define several textual expressions that describe different attribute states (\textit{e.g.}, ``old man'', ``very old person'', ``young man'', ``very young person'').

To enrich visual diversity and reduce identity bias, we prepend each prompt with descriptive prefixes drawn from a fixed pool. These include phrases like ``a photo of a'', ``a photo of a beautiful'', as well as identity-related descriptors such as ``a photo of a black / white / asian / arab / caucasian'', and their compositional forms (\textit{e.g.}, ``a photo of a beautiful asian'').

We construct the full prompt set by exhaustively combining all prefixes and prompt templates for each attribute. If the total number of combinations is less than 1000, we retain all; otherwise, we randomly sample 1000 combinations using different random seeds to ensure diversity across prompt sets. This results in a balanced and diverse prompt distribution per attribute.

\paragraph{Attribute List.} We consider a total of 52 facial attributes and use their corresponding text embeddings to train the Attribute Sparse Autoencoder. These attributes span appearance, expression, hairstyle, and accessory-related features. The full list is as follows: \textit{age, angry, arched-eyebrows, bags-under-eyes, bald, bangs, big-lips, big-nose, birthmarks,
black-hair, blond-hair, brown-hair, bushy-eyebrows, chubby, colorful-outfit, double-chin,
elegant, eye-size, eyeglasses, fitness, freckled, frowning, gray-hair, groomed, high-cheekbones,
makeup, moles, mouth-slightly-open, multicolored-hair, mustache, narrow-eyes, oval-face, pale-skin,
pierced, ponytail, pouting, receding-hairline, rosy-cheeks, sad, scarred, sideburns, smile,
surprised, tattooed, tiredness, wavy-hair, wearing-earrings, wearing-hat, wearing-lipstick,
wearing-necklace, wearing-necktie, width.}

\paragraph{Image Editing Evaluation Criteria.}
In the evaluation of image edits, we introduce Qwen Score (QS) to assess semantic attribute changes or facial expression. This is done using a 5-point scale across three dimensions: Action / Expression Fidelity, Identity Preservation, and Visual \& Anatomical Coherence. This system is based on the scoring framework provided by the Qwen2.5-VL-Instruct model. The criteria are as follows:

\textbf{Expression Fidelity} captures how faithfully the edited image reflects the target expression described in the instruction. A high score reflects a precise match in terms of intensity and orientation. In contrast, lower scores are triggered by absent or incorrectly rendered expressions, partial changes (e.g., only some facial muscles involved), or misaligned pose dynamics.

\textbf{Identity Preservation} evaluates whether the subject remains recognisable and consistent across the original and edited images. Perfect preservation includes consistent facial structure, hairstyle, and clothing. Score degradation occurs when noticeable drifts—such as altered facial features, skin tone, or missing accessories—make the subject appear different or less identifiable.

\textbf{Visual \& Anatomical Coherence} focuses on the realism and structural consistency of the rendered image. High scores reflect anatomically correct bodies, natural lighting and shadows, and coherent textures. Lower scores are given in the presence of visual artifacts, anatomical distortions, or rendering inconsistencies such as cut-out edges, lighting mismatch, or unrealistic joint positions. 

\begin{figure}[!h]
    \centering
    \includegraphics[width=0.70\linewidth]{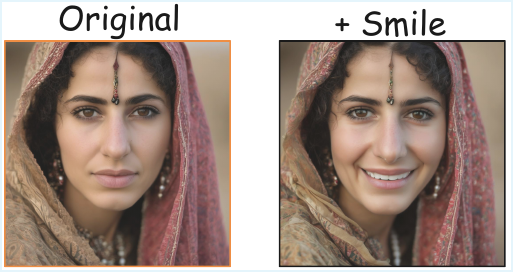} 
    \caption{
        Attribute manipulation results for \textit{Smile}. The image shows the subject before and after the expression modification.} 
    \label{fig:smile-score} 
\end{figure}

Each image pair (before and after steering) is evaluated in terms of three dimensions: Expression Fidelity, Identity Preservation, and Visual \& Anatomical Coherence, each scored on a 5-point scale. In addition to the numerical scores, a brief reasoning (within 20 words) is provided to justify the evaluation results. As an example, we present the evaluation of an manipulated image, for which the Qwen model assigned scores of 4 for Action Fidelity, 5 for Identity Preservation, and 4 for Visual \& Anatomical Coherence (see Figure~\ref{fig:smile-score}). The manipulated image shows a successful transition from a neutral expression to a clear, natural-looking smile, while preserving consistency in the head pose and facial features. Minor inaccuracies in the smile’s angle or intensity lead to an Action Fidelity score of 4. The subject’s identity is perfectly preserved, justifying a score of 5 for Identity Preservation. The image also maintains visual and anatomical coherence, with only slight imperfections, resulting in a score of 4 for Visual \& Anatomical Coherence. According to Qwen’s analysis, the brief explanation for this result is: \textit{``The subject's expression has changed to a smile, but the overall pose remains the same"}.

\paragraph{Multi-Subject Manipulation.}
The goal of multi-subject manipulation is to steer the facial attributes of a specific target subject (\textit{e.g.}, a man or a woman) within a multi-subject scene, while preserving the other subjects and the rest of the scene consistency. As shown in Sec.4.5 in the main paper, we employ the fine-tuning objective $\mathcal{L}_{\text{multi}} = \mathcal{L}_{\text{sae}} + \eta \mathcal{L}_{\text{cons}}$, where $\eta$ is set to 0.1. 
During inference, we first encode the paired sentences hidden states $\text{H}^+$ and $\text{H}^-$ (with and without the target attribute) into their corresponding sparse latent representations, denoted as $z^+$ and $z^-$, using the pre-trained Attribute Sparse Autoencoder. These representations are then passed through the Attention Pooling (AttPooling) Aggregator (AAg) module, which aggregates the attribute-related semantics into compact token-level embeddings. The aggregated directional vector, $\Delta z = \text{AAg}(z^+) - \text{AAg}(z^-)$, captures the pure attribute semantics. This vector is then decoded via the SAE decoder and added to the target subject's embedding $e_{\text{target}}$, yielding the updated subject embedding: $e_{\text{target}}' = e_{\text{target}} + \text{DEC}(\lambda \cdot \Delta z)$.
This process ensures that the attribute manipulation is precisely localized to the target subject, while preserving the identity and context of other elements in the scene.

\paragraph{Photography Style List.} We train our model on a diverse set of photography styles, including: \textit{architectural, astrophotography, black-and-white, blue hour, bokeh, cinematic, documentary, dreamy, fashion, film grain, fisheye, food, golden hour, HDR, high contrast, infrared, landscape, long exposure, low light, macro, minimalist composition, monochrome, muted colors, natural lighting, neon lighting, overexposed, pastel tone, portrait, product, sepia tone, soft focus, street photography, studio lighting, surreal, telephoto, underexposed, vintage, wide-angle}.

\section{Additional Experimental Results}
\subsection{Additional attribute Manipulation}
We present additional qualitative results demonstrating our method’s ability to manipulate a wide range of attributes in an identity-preserving manner, as illustrated in Figures~\ref{fig:app-continue} to~\ref{fig:app-multi}.
Each figure below shows a series of images in rows, where the leftmost column is the original input image, and the subsequent columns represent progressive edits along a specific semantic attribute. 

\subsection{Training SAE with FLUX}
\begin{figure}[!h]
    \centering
    \includegraphics[width=0.70\linewidth]{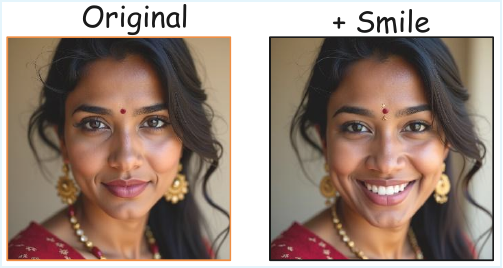} 
    \caption{ 
     Smile manipulation on Flux using our method. The generated results maintain the original identity and scene context while achieving realistic smile steering. 
    }
    \label{fig:app-flux} 
\end{figure}

To evaluate the generalization of our method, we train the SAE on the FLUX.1-dev~\cite{flux2024} model. Text embeddings are extracted from the 23rd layer of the T5 text encoder (out of 24 layers in total). Qualitative results of this experiment are presented in Figure~\ref{fig:app-flux}.

\subsection{Comparison with Raw Embeddings}
We compare our sparse direction, $W_{\text{dec}}(\lambda \times \text{ENC}(x_\text{A}))$, against raw embedding addition, $\lambda \times x_\text{A}$, in Table~\ref{tab:ablation-raw-emb}. Our method consistently outperforms $\text{Emb}_{\text{raw}}$ across all attributes and metrics, yielding an average improvement of 0.212 in QS and 0.196 in IS. This confirms that SAE-derived directions capture purer semantic signals than the raw embeddings from the pretrained text encoder. 
As noted in~\cite{NEURIPS2024_996bef37}, original embedding spaces are often dense and semantically entangled, which can cause unintended identity shifts (see Fig.1 in the main paper). By mapping these into a high-dimensional sparse space, our approach achieves semantic disentanglement by decomposing representations into their underlying, interpretable components. This alignment with SpLICE~\cite{NEURIPS2024_996bef37} further demonstrates the effectiveness of decomposing dense representations into sparse, clean units.

\begin{table}[!h]
\caption{Comparison with raw attribute embedding.}
\vspace{-2mm}
\centering
\resizebox{1.0\columnwidth}{!}{
\begin{tabular}{c|cc|cc|cc|cc}
\toprule
\multirow{2}{*}{Method} & \multicolumn{2}{c|}{\textbf{Old}} & \multicolumn{2}{c|}{\textbf{Smile}} & \multicolumn{2}{c|}{\textbf{Makeup}} & \multicolumn{2}{c}{\textbf{Avg}} \\
& QS & IS & QS & IS & QS & IS & QS $\uparrow$ & IS $\uparrow$ \\
\midrule
Emb$_\text{raw}$  &3.791  &0.497  &4.220 &0.520  &3.960  &0.489  &3.990  &0.502  \\
Ours  & \textbf{4.049} & \textbf{0.716} &\textbf{4.265} &\textbf{0.637} &\textbf{4.291} &\textbf{0.742} &\textbf{4.202} &\textbf{0.698} \\
\bottomrule
\end{tabular}
\label{tab:ablation-raw-emb}
}
\end{table}

\begin{figure}[h]
    \centering
    \includegraphics[width=1.0\linewidth]{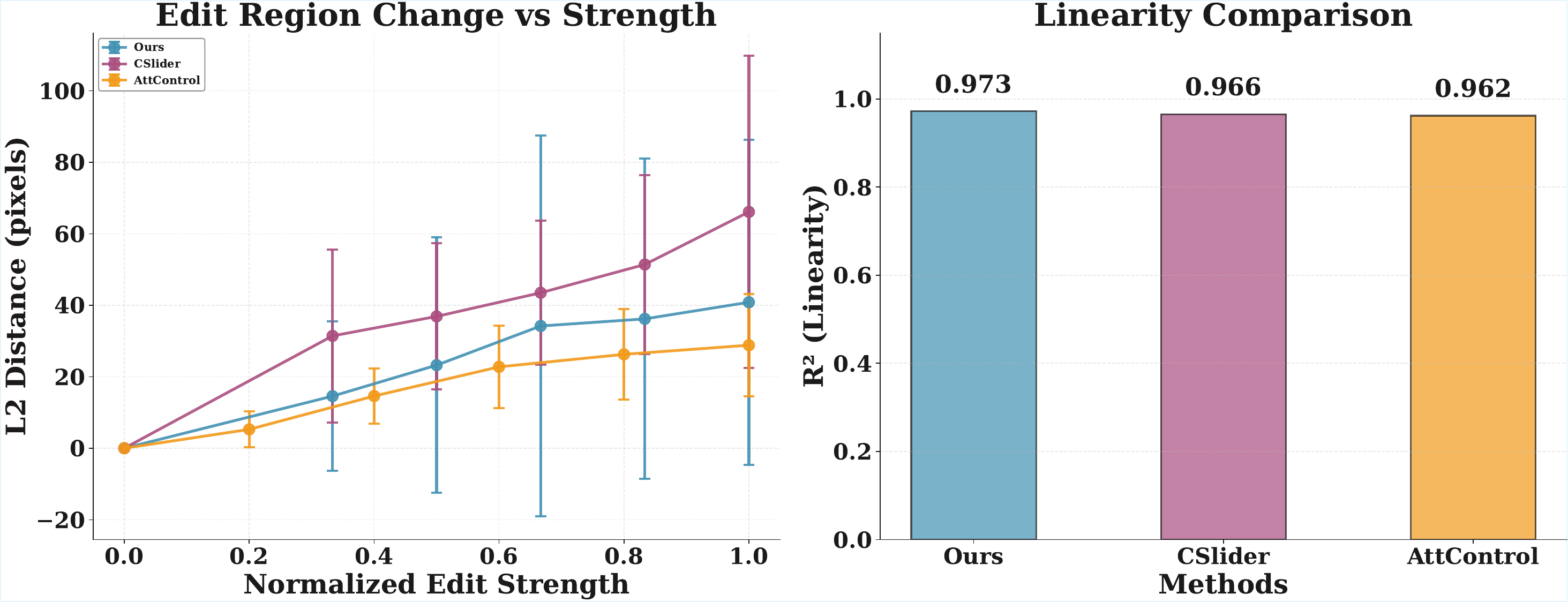} \\
    \vspace{-1mm}
    \caption{Geometric edit continuity and linearity comparison.}
    \vspace{-4mm}
    \label{fig:linear} 
\end{figure}
\subsection{Geometric metrics for continuity}
We adopt dlib landmarks to track facial geometry changes, computing L2 pixel distances within the edit region across steering strengths. As shown in Figure~\ref{fig:linear}, all methods achieve monotonic geometric progression, with our method exhibiting a moderate edit magnitude (neither over-editing like \textit{CSlider} nor under-editing like \textit{AttControl}). More importantly, our method achieves the highest linearity ($R^2$ of 0.973), compared to \textit{CSlider} (0.966) and \textit{AttControl} (0.962), confirming its superior precision in continuous control.

\subsection{Visualize sparsity of latent space}
\begin{figure}[h]
    \centering
    \includegraphics[width=1.0\linewidth]{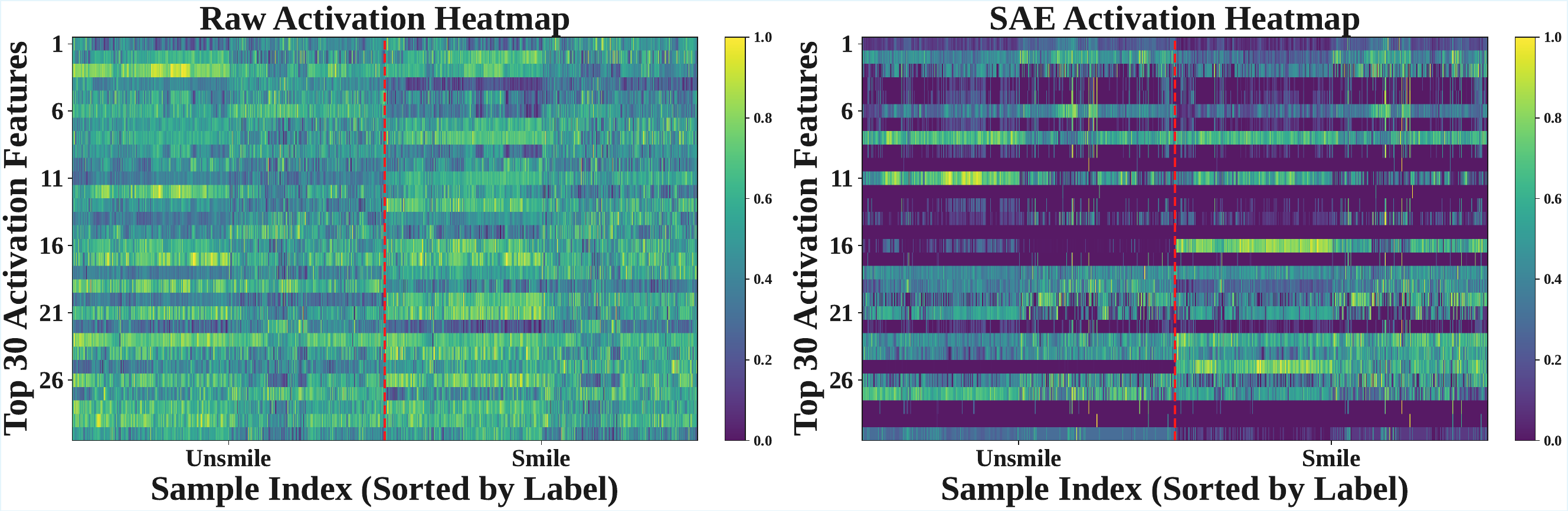} \\
    \vspace{-1mm}
    \caption{ Raw vs. SAE latent space sparsity.}
    \vspace{-4mm}
    \label{fig:activation_heatmap} 
\end{figure}

We compare raw activations (\textit{i.e.}, CLIP text embeddings) with those in our SAE latent space. As shown in Figure~\ref{fig:activation_heatmap}, raw embeddings are semantically entangled, whereas our SAE yields a sparse, axis-aligned representation. In this space, distinct dimensions are selectively triggered for specific attributes (\textit{Smile} vs.\textit{Unsmile}) samples.

\section{Prompt Information}
For reproducibility, we provide the text prompts used to generate the images in the appendix figures. 
\subsection{Prompts for Main Figures}
The text prompts used to generate the images for the main figures are as follows.\\
Figure 1: (1) \textit{A photo of a gentle Thai man with layered hair, in a warm-toned studio.}\\ 
Figure 1: (2) \textit{A photo of an honest Dutch man in a white button-up shirt, with a blurred background.}\\ 
Figure 1: (3) \textit{A high-quality portrait of a man in a tuxedo, with clear, detailed facial features. } \\ 
Figure 4: Prompt: \textit{A close-up of a woman wearing a red scarf, looking thoughtful.}\\ 
Figure 5: \textit{A photo of an honest Latin American talented woman. } \\ 
Figure 6: (1) \textit{A woman with medium-length blonde hair, pale skin, and soft blue eyes.}\\ 
Figure 6: (2) \textit{A photo of a peaceful Middle Eastern woman.}\\  
Figure 7: \textit{A photo of a woman wearing a red scarf.}\\ 
Figure 8: (1) \textit{A man wearing glasses while holding a banana.}\\
Figure 8: (2) \textit{A man in front of a Christmas tree with his dog.} \\
Figure 9: (1) \textit{A photo of a forest.} \\ 
Figure 9: (2) \textit{A path through a dense forest. } \\ 
Figure 10: (1) \textit{A portrait of a woman and a man, standing under cherry blossoms.}\\ 
Figure 10: (2) \textit{A portrait of a woman and a man holding cameras in a motorcycle garage with subtle tones.}\\ 
Figure 10: (3) \textit{A portrait of a woman and a man, wearing scarves in a winter city, in gentle morning sunshine. } \\  
Figure 10: (4) \textit{A portrait of a woman and a man wearing photography gear at a music studio in a cool color palette.}  

\subsection{Prompts for Appendix Figures}
The text prompts used to generate the images in the appendix are listed below.\\
Figure~\ref{fig:app-continue} prompts:\\
\textit{Smile}:
(1) \textit{A photo of a charming Singaporean man with clean-cut hair, with studio backdrop.}\\  
(2) \textit{A photo of a composed Moroccan man wearing a woolen scarf, with a blurred background. }\\  
(3) \textit{A photo of an honest Dutch man in a white button-up shirt, with a blurred background.} \\
\textit{Old}:
(1) \textit{A photo of a refined Italian man wearing a checkered blazer, with soft window light. }\\  
(2) \textit{A photo of a serene Danish man with round glasses, near a sunlit window. }\\  
(3) \textit{A photo of a poised Greek person in a linen shirt, in a studio with soft shadows.} \\
\textit{Wearing-Lipstick}:
(1) \textit{A photo of an honest Latin American talented woman.}\\
(2) \textit{A woman with short hair, looking at the camera.}\\
\textit{Afro}: \textit{A portrait of a woman standing by the beach at sunset, with soft and natural lighting. }\\  
\textit{Wavy-Hair}: \textit{A photo of a composed Swedish man with short blond hair, with a blurred background.}\\ 

Figure~\ref{fig:app-att} prompts:\\
\textit{Eyeglasses}: (1) \textit{A photo of an elegant Chinese man wearing a red scarf, with a blurred background.}\\  
\textit{Eyeglasses}: (2) \textit{A photo of a cheerful Filipino man wearing a red scarf, with a soft gray backdrop.}\\ 
\textit{Hat}: (1) \textit{A photo of a beautiful woman.}\\ 
\textit{Hat}: (2) \textit{A photo of a wise European man wearing a casual sweater, in a sunny environment.}\\ 
\textit{Bald}: (1)\textit{ A photo of a gentle Vietnamese man with short cropped hair, with a soft gray backdrop.} \\ 
\textit{Bald}: (2) \textit{A photo of a serene Danish man with round glasses, near a sunlit window. }\\ 
\textit{Blond-Hair}: (1) \textit{A photo of a sophisticated Russian woman with a pearl necklace, in a minimalist room.}\\ 
\textit{Blond-Hair}: (2) \textit{A woman with short hair, looking at the camera.} \\ 


Figure~\ref{fig:app-style} photography style prompts:\\
(1) \textit{A photo of sun rays through tall trees.} \\
(2) \textit{A peaceful scene of a lake at sunset.} \\
(3) \& (6) \textit{A group of tall trees in a park. }\\
(4) \& (5) \textit{A photo of a lone tree in a vast field.} \\
(7) \textit{A sunrise over a calm ocean.} \\
(8) \textit{A photo of a snowy forest path.} \\

Figure~\ref{fig:app-multi} prompts:\\
(1) \textit{A portrait of a woman and a man framing a shot together in a snowy village in cinematic lighting.}\\
(2) \textit{A close-up of a woman and a man, front of a cafe, with crisp autumn light.}\\
(3) \textit{A portrait of a woman and a man, wearing scarves in a winter city, in gentle morning sunshine.}\\
(4) \textit{ A portrait of a woman and a man standing by the beach at sunset, with soft and natural lighting.}\\



%

\begin{figure*}[!h]
    \centering
    \includegraphics[width=1.0\linewidth]{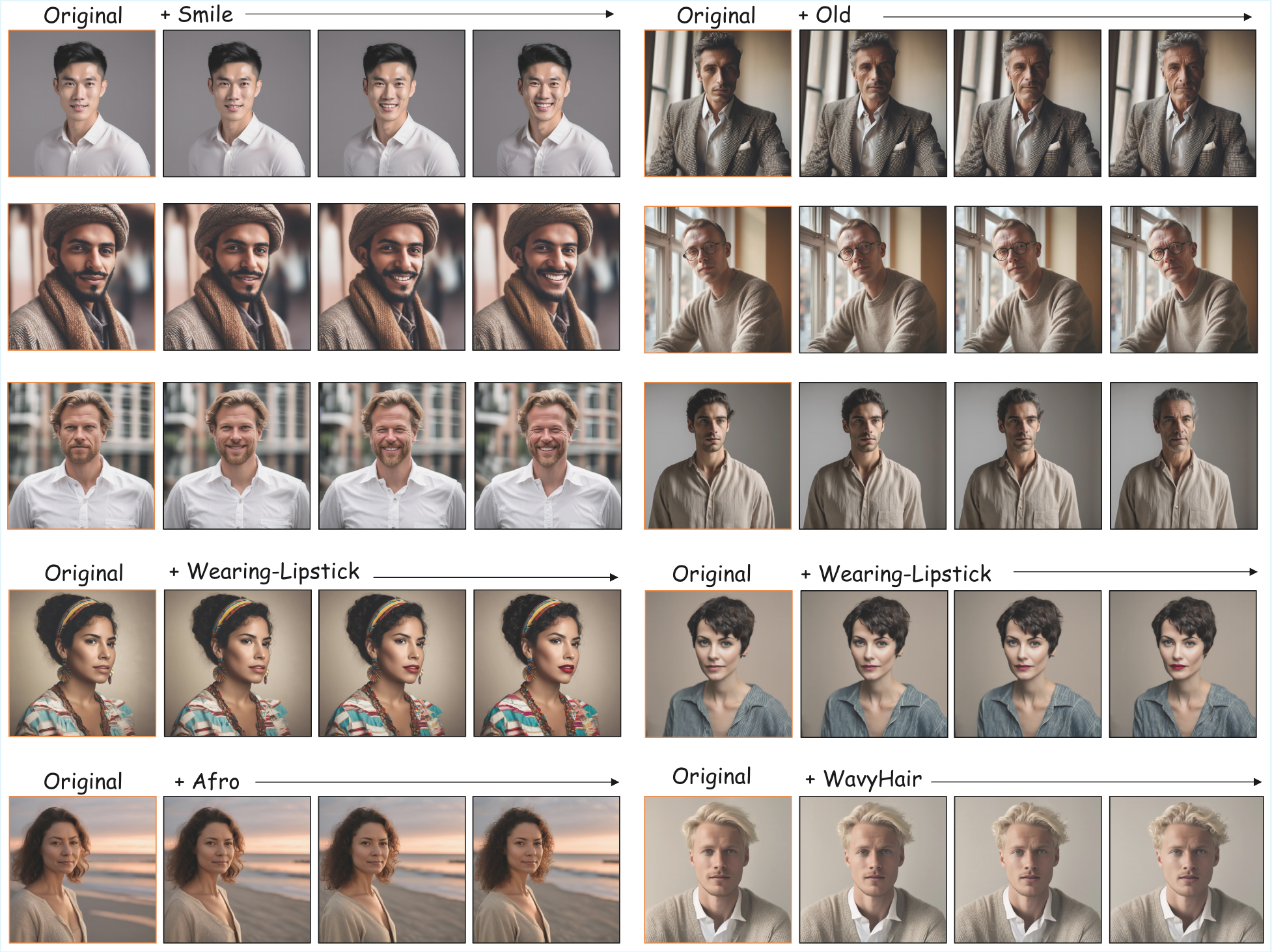} 
    \caption{
    Continuous manipulation of the \textit{Smile}, \textit{Old}, \textit{Wearing-Lipstick}, \textit{Afro} and \textit{WavyHair} attribute. Each row shows a subject transitioning from neutral to the attribute steered, with identity and lighting well preserved.
    }
    \label{fig:app-continue} 
\end{figure*}

\begin{figure*}[!h]
    \centering
    \includegraphics[width=1.0\linewidth]{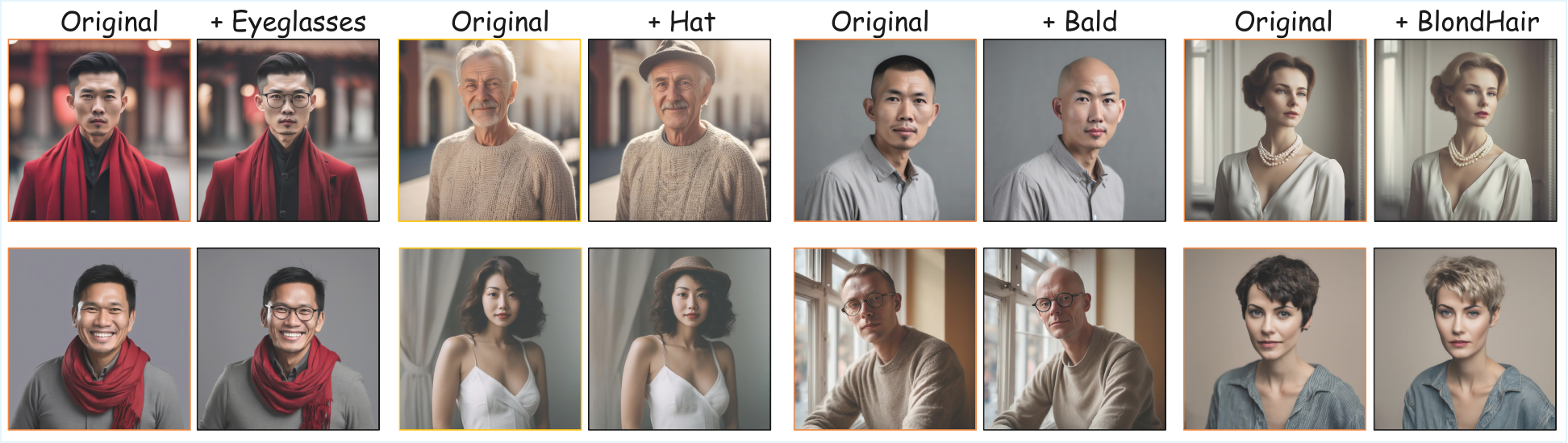} 
    \caption{
    Attribute manipulation results across various appearance traits. Each pair shows the subject before (left) and after (right) editing, demonstrating the effect of adding attributes, such as \textit{Eyeglasses}, \textit{Hat}, \textit{Bald}, and \textit{BlondHair}. The results illustrate the model's ability to generate realistic and identity-consistent modifications across a diverse set of visual changes.
    }
    \label{fig:app-att} 
\end{figure*}


\begin{figure*}[!t]
    \centering
    \includegraphics[width=1\linewidth]{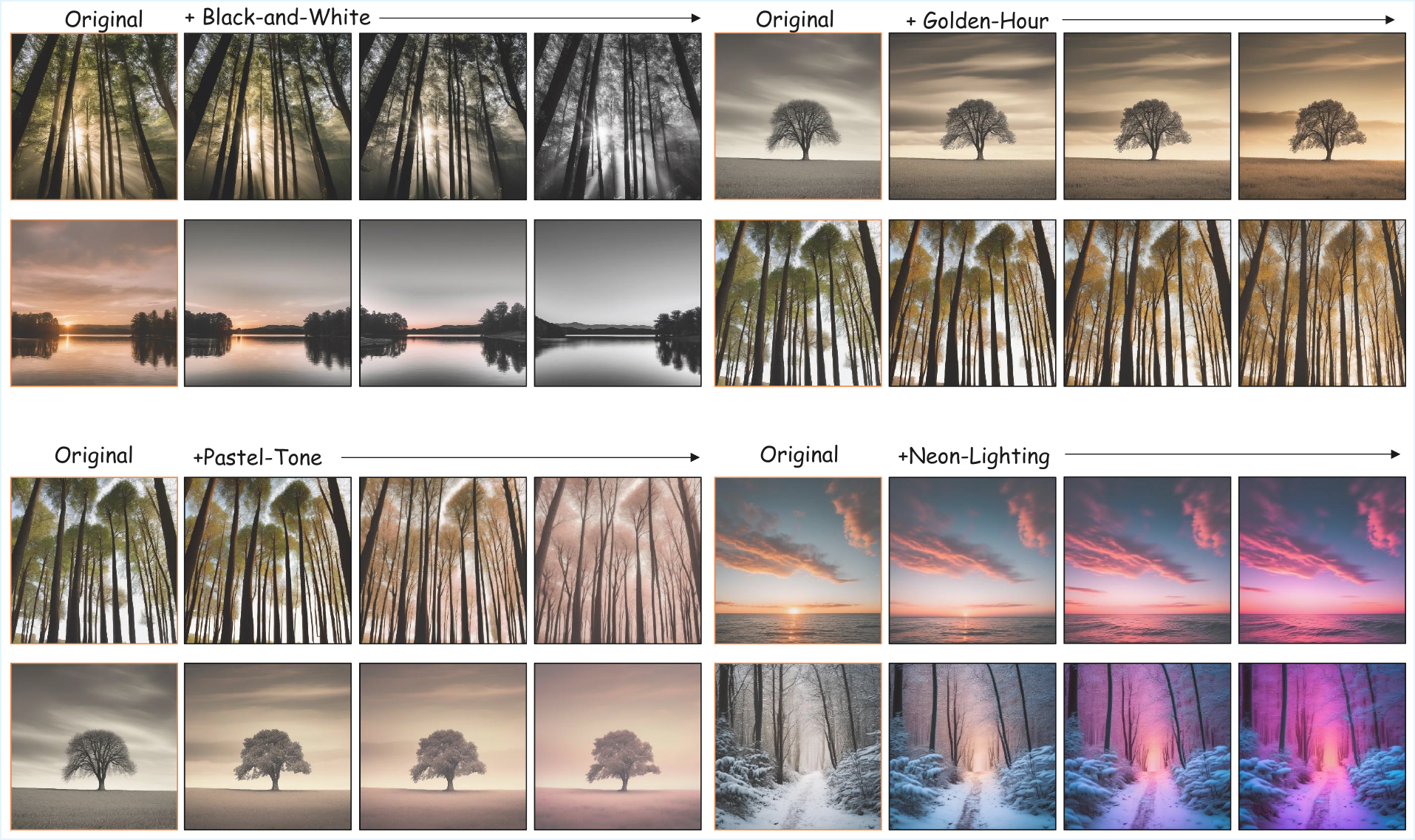} 
    \caption{
    Continuous manipulation of  photography style, such as the \textit{Black-and-White}, \textit{Golden-Hour}, \textit{Pastel-Tone} and \textit{Neon-Lighting} attribute. Each row demonstrates the style transition on the original images while maintaining their core structural consistency.
    }
    \label{fig:app-style} 
\end{figure*}

\begin{figure*}[!t]
    \centering
    \includegraphics[width=1\linewidth]{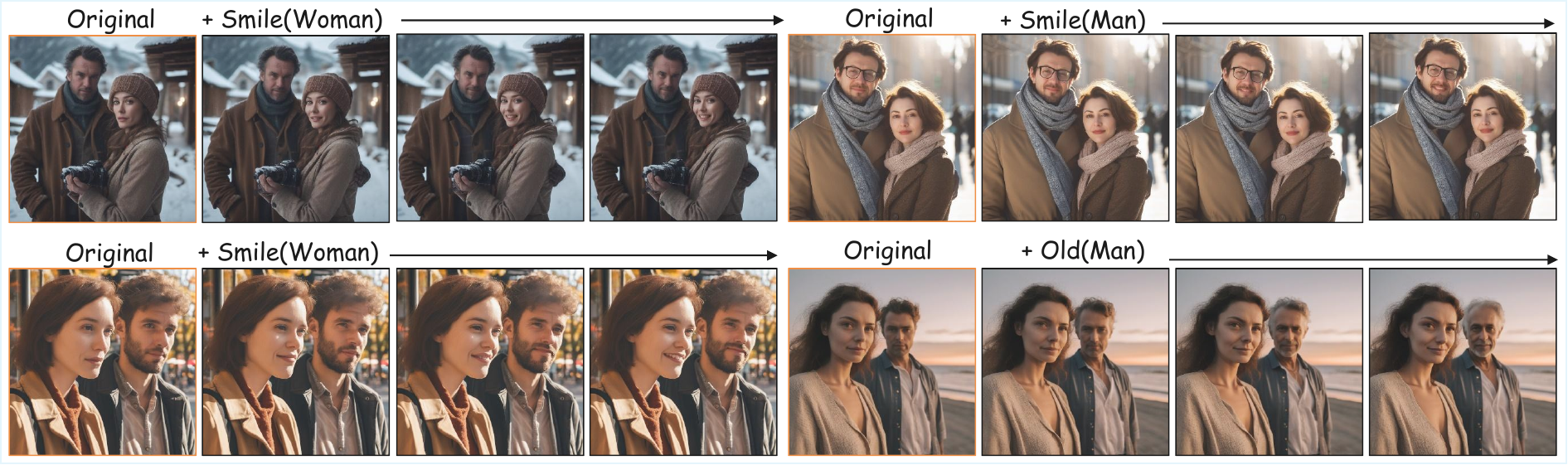} 
    \caption{
    Continuous manipulation of  attribute in a multi-subject scene. Each row demonstrates the continuous application of an attribute (\textit{e.g.}, \textit{Smile}, \textit{Old}) to a specific subject (\textit{e.g.}, \textit{man} or \textit{woman}) while keeping the other subject unchanged.
    }
    \label{fig:app-multi} 
\end{figure*}

\end{document}